\documentclass{esannV2}

\usepackage{graphicx}
\usepackage{hyperref}
\usepackage{acronym}
\usepackage[utf8]{inputenc}
\usepackage[numbers,sort]{natbib}
\usepackage{pifont}
\usepackage{amsmath}
\usepackage{float}
\usepackage{tikz}
\usepackage{rotating}
\usepackage{subcaption}
\usepackage{booktabs}
\usepackage{pgfplots}
\pgfplotsset{compat=1.14}
\usepackage{verbatim}
\usepackage{amsthm}
\usepackage{tikz}
\usepackage[capitalise]{cleveref}
\usepackage[utf8]{inputenc}
\usepackage{amssymb,amsmath,array}
\usepackage{xspace}
\usepackage{xargs}                      % Use more than one optional parameter in a new commands
\usepackage{listings}
\usepackage
[colorinlistoftodos,
prependcaption,
textsize=small,
textwidth=4cm]
{todonotes}

\theoremstyle{definition}

\theoremstyle{remark}

\crefname{section}{section.}{section}
\crefname{figure}{Figure}{Figures}
\Crefname{figure}{Figure}{Figures}
\crefname{section}{Sect.}{Sect.}
\crefname{definition}{Def.}{Def.}
\crefname{Definition}{Def.}{Def.}
\Crefname{section}{Section}{Sections}

\usetikzlibrary{arrows,shapes,automata,positioning,calc,shapes.symbols,shapes.geometric,shadows}
\usetikzlibrary{decorations.pathmorphing,decorations.pathreplacing,decorations.shapes,chains}

\renewcommand{\comment}[1]{}
\newcommand{\RNum}[1]{\uppercase\expandafter{\romannumeral #1\relax}}

\newcommandx{\me}[2][1=]{\todo[linecolor=cyan,backgroundcolor=cyan!25,bordercolor=cyan,#1]{\textbf{ME: }#2}\xspace}

\newcommandx{\unsure}[2][1=]{\todo[linecolor=red,backgroundcolor=red!25,bordercolor=red,#1]{#2}\xspace}
\newcommandx{\change}[2][1=]{\todo[linecolor=blue,backgroundcolor=blue!25,bordercolor=blue,#1]{#2}\xspace}
\newcommandx{\info}[2][1=]{\todo[linecolor=green,backgroundcolor=green!25,bordercolor=green,#1]{#2}\xspace}
\newcommandx{\improvement}[2][1=]{\todo[linecolor=Plum,backgroundcolor=Plum!25,bordercolor=Plum,#1]{#2}\xspace}
\newcommandx{\thiswillnotshow}[2][1=]{\todo[disable,#1]{#2}\xspace}

\newacro{PGP}{Predictive Gating Pyramid}
\newacro{Conv-PGP}{Convolutional Predictive Gating Pyramid}
\newacro{CConv-PGP}{Content Convolutional Predictive Gating Pyramid}
\newacro{GAE}{Gated AutoEncoder}
\newacro{DNN}{Deep Neural Network}
\newacro{VLN}{Video Ladder Network}
\newacro{VPN}{Video Pixel Network}
\newacro{MM}{Moving MNIST}
\newacro{BB}{Bouncing Ball}
\newacro{OMM}{Occluded Moving MNIST}
\newacro{RAE}{Relational Autoencoder}

\usepackage{placeins}
\usepackage{float}
\usepackage{eso-pic}

\AtBeginDocument{\AddToShipoutPictureFG*{\AtTextUpperLeft{\put(0,\LenToUnit{40pt}){\parbox{\textwidth}{\centering\bfseries
                                        27th European Symposium on Artificial Neural Networks, Computational Intelligence and Machine Learning (ESANN), Bruges, Belgium, 2019
}}}}}

\begin{document}

%\listoftodos[Comments]
\newpage
\title{Complex Valued Gated Auto-encoder \\for Video Frame Prediction}

%***********************************************************************
% AUTHORS INFORMATION AREA
%***********************************************************************
\author{
Niloofar Azizi, Nils Wandel, Sven Behnke
%
% Optional short acknowledgment: remove next line if non-needed
%\thanks{}
%
% DO NOT MODIFY THE FOLLOWING '\vspace' ARGUMENT
\vspace{.1cm}\\
%
% Addresses and institutions (remove "1- " in case of a single institution)
\textsuperscript{}({azizi,wandeln,behnke}@ais.uni-bonn.de)
\vspace{.1cm}\\
\textsuperscript{}Bonn University, Computer Science Department,\\
Endenicher Allee 19a, 53115 Bonn, Germany
%
% Remove the next three lines in case of a single institution
\vspace{.1cm}\\
}
%***********************************************************************
% END OF AUTHORS INFORMATION AREA
%***********************************************************************

\maketitle

\begin{abstract}
\footnotesize
In recent years, complex valued artificial neural networks have gained increasing interest as they allow neural networks to learn richer representations while potentially incorporating less parameters. Especially in the domain of computer graphics, many traditional operations \comment{such as image smoothing / sharpening} rely heavily on computations in the complex domain, thus complex valued neural networks apply naturally.

In this paper, we perform frame predictions in video sequences using a complex valued gated auto-encoder. First, our method is motivated showing how the Fourier transform can be seen as the basis for translational operations. Then, we present how a complex neural network can learn such transformations and compare its performance and parameter efficiency to a real-valued gated auto-encoder. Furthermore, we show how extending both --- the real and the complex valued --- neural networks by using convolutional units can significantly improve prediction performance and parameter efficiency.

The networks are assessed on a moving noise and a bouncing ball dataset.

\end{abstract}

\section{Introduction}
\label{introduction}

Video prediction is the task of predicting future frames by extracting complex spatio-temporal features from a sequence of seed frames. In recent years \acfp{DNN} showed promising results in video prediction \cite{xingjian2015convolutional,mathieu2015deep}.

\citet{michalski2014modeling} proposed the \ac{PGP} architecture to learn and predict the transformation in a sequence of frames. In \ac{PGP} as well as its equivalent fully convolutional architecture \cite{demodeling}, a layer of mapping units encodes transformation using a \ac{GAE}. The \ac{GAE} is designed based on the assumption that two temporally consecutive frames can be interpreted as a linear transformation of one another. \ac{GAE} was improved by \citet{droniou2013gated} by going into complex domain. Recently the analysis of \ac{DNN} architectures in complex domain raised attention as it makes the learning process faster \cite{arjovsky2016unitary} and the optimization process easier \cite{nitta2002critical}.

In this paper, we extend the \ac{GAE} with tied input weights \cite{droniou2013gated} to perform video frame prediction and propose a convolutional form which drastically reduces the number of model parameters while significantly improving the performance on a Bouncing Balls dataset.

\comment{

\begin{figure*}[tb]
\centering
\begin{subfigure}{\columnwidth}
  \centering
  \input{figures/architecture/Conv-PGP-tied.tex}
\label{fig:arch:convpgp}
\vspace{3mm}
\end{subfigure}
\caption{One layer Tied-Conv-PGP architecture.}
\label{fig:arch}
\end{figure*}
}
\section{Frame Prediction using Deconvolution}
\label{motivating_example}

A motivating example shows how the transformation between two images that are translated copies can be calculated by deconvolution.

Let $X_{t-1}$ be the first image and $X_t(x,y)=X_{t-1}(x-t_x,y-t_y)$ be the second image corresponding to $X_{t-1}$ translated by $t_x,t_y$. Then, the transformation of $X_{t-1} \Rightarrow X_t$ can be seen as a convolution of $X_{t-1}$ with a $\delta$-function $\delta(x-t_x,y-t_y)$:
\begin{equation}
    X_t(x,y) = \int X_{t-1}(\hat{x},\hat{y}) \delta(x-t_x-\hat{x},y-t_y-\hat{y}) d\hat{x}d\hat{y} = X_{t-1}(x-t_x,y-t_y)
\end{equation}

Thus in order to obtain $\delta(x,y)$, one can deconvolve $X_t$ with $X_{t-1}$:
\begin{equation}
    \delta(x-t_x,y-t_y) = \left(\mathcal{F}^{-1}\frac{(\mathcal{F}X_t)(u,v)}{(\mathcal{F}X_{t-1})(u,v)}\right)(x,y)
\end{equation}

Here, $\mathcal{F}$ denotes the two dimensional Fourier transform and $\mathcal{F}^{-1}$ denotes the two dimensional inverse Fourier transform.
%\begin{align}
%    (\mathcal{F} X)(u,v) &= \iint X(x,y) e^{-i(ux+vy)} dx dy\\
%    (\mathcal{F}^{-1} U)(x,y) &= \iint U(u,v) e^{i(ux+vy)} du dv
%\end{align}

After having obtained the transformation $\delta$, it can be used to extrapolate $X_t$ and calculate $X_{t+k}$:
\begin{equation}
    X_{t+k} = \mathcal{F}^{-1} \left( (\mathcal{F}\delta)^k \cdot \mathcal{F}X_t  \right) = \mathcal{F}^{-1} \left( \left(\frac{(\mathcal{F}X_t)(u,v)}{(\mathcal{F}X_{t-1})(u,v)}\right)^k \cdot \mathcal{F}X_t \right)
\end{equation}
A schematic depiction of these operations can be found in Figure \ref{model_architectures} a).
While this works in theory, in practice usually several problems occur: the problem is ill posed, thus very sensitive to noise and in principle even multiple solutions could be obtained. $|\mathcal{F}X_{t-1}|$ can become arbitrarily small so one usually has to add a tiny offset $\epsilon$ in the denominator. Furthermore, this method makes the assumption of a periodic boundary and a uniform translation of the whole image. 

Thus we want to train a model which can learn by itself more robust basis transformations that are not only suited for translations but can also handle for example rotations. Also, the method should be able to cope with multiple different local transformations arising from different objects in the scene.

\section{Gated Auto-encoders}

\begin{figure}[h]
\centering
\includegraphics[width=1\linewidth]{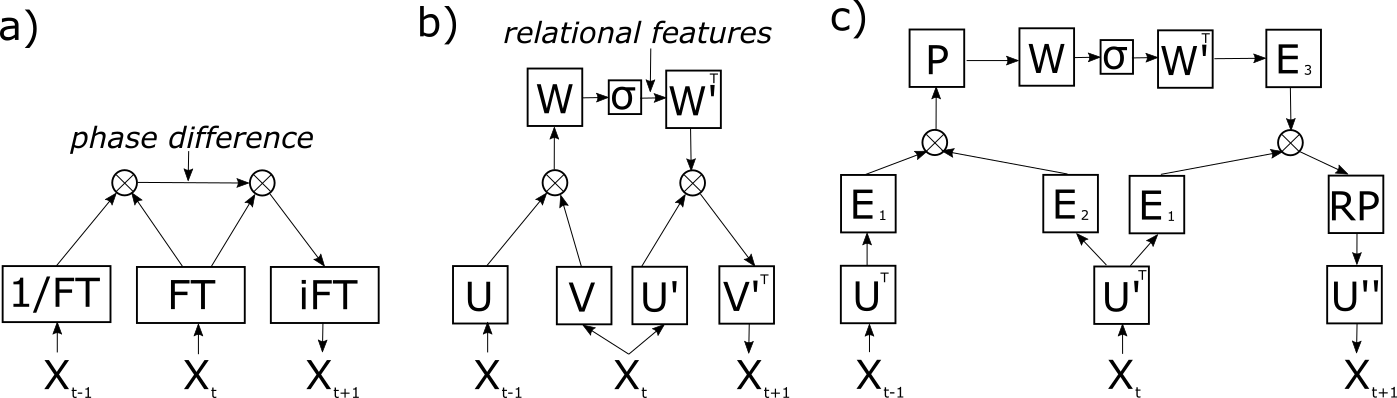}
\caption{a) Frame prediction using deconvolution, b) real valued GAE, c) complex valued GAE. $U,U',U'',V,V',W,W'$ denote, depending on the version of the architecture, either fully connected or convolutional units. The weights of the real and complex valued GAE are shared among $(U,U',U''),(V,V')$. Sharing weights among $(W,W')$ strongly decreased performance.}
\label{model_architectures}
\centering
\end{figure}

We refer to Gated Auto-Encoders \cite{memisevic2013learning} as "real valued GAE" and to Gated Auto-Encoders with tied input weights  \cite{droniou2013gated} as "complex valued GAE". If the transformation between images is linear, it can be written as:
\begin{equation}
    X_t = L X_{t-1}
\label{eq:1}
\end{equation}
Here, $X_t$ and $X_{t-1}$ denote the vectorized form of the two images and $L$ denotes a transformation matrix.
If we further assume, that the transformation is orthogonal, $L$ can be decomposed into:
\begin{equation}
    L = UDU^*
\label{eq:2}
\end{equation}
with U being a unitary matrix ($UU^*=I$) and D being a diagonal matrix containing complex numbers of absolute value 1. While the assumption of a linear orthogonal transformation at first glance seems quite restricting it still comprises for example all transformations that can be described as pixel permutations (e.g. translation / rotation / shearing). From \ref{eq:1} and \ref{eq:2}, it follows that:
\begin{equation}
    U^*X_t = D U^* X_{t-1} \Rightarrow D = \mathop{\mathrm{diag}}\left( \frac{U^* X_t}{U^*X_{t-1}}\right)
\label{eq:3}
\end{equation}
This is remarkable since it shows how orthogonal linear transformations can be represented in a much more compact way as rotations in the complex plane by $D$ when the basis is properly changed by $U$. If we consider, for example, only translational transformations, $U^*$ basically becomes a discrete Fourier transform and $D$ corresponds to phase-differences in frequency domain. This corresponds exactly to what is described in Section \ref{motivating_example} and again, the obtained representation of the transformation can be used to extrapolate to future frames:
\begin{equation}
    X_{t+k} = U D^k U^* X_t
\label{eq:4}
\end{equation}
As for the introductory example, this leads to problems when projections of $U$ lead to small absolute values, because in this case the computation of $D$ becomes ill-conditioned resulting in falsely detected transformations.

Thus, the real valued GAE \cite{memisevic2013learning} as depicted in Figure \ref{model_architectures} b) was developed. In this architecture, two separate trainable linear modules $U$ and $V$ learn representations of $U^*$ and phase-shifted representations of $U^*$, respectively. Furthermore, in order to properly normalize the transformation representation, two additional linear modules ($W$ and $W'^T$) and a sigmoid activation function are used. This representation then is either multiplied ("gated") by $U' X_{t-1}$ in the case of reconstruction or by $U' X_t$ in the case of prediction and projected back by $V'^T$:
\begin{align}
    X_t &= V'^T (U'X_{t-1} \cdot W'^T\sigma(W(U X_{t-1}\cdot V X_t))) &\mathrm{reconstruction}\\
    X_{t+1} &= V'^T (U'X_t \cdot W'^T\sigma(W(U X_{t-1}\cdot V X_t))) &\mathrm{prediction}
\end{align}
This real valued GAE is able to learn robust relational features for a wide range of linear transformations \cite{memisevic2013learning}. However, \cite{droniou2013gated} pointed out that parameter efficiency can be drastically increased since $U$ and $V$ learn mostly the same features. A complex valued GAE was suggested, which directly makes use of Eq. \ref{eq:3}. By treating $U^*$ in the complex domain, it is not further needed to learn $U$ and $V$ separately. Instead, by carefully designing the matrices $E_1,E_2,E_3,P,R$ (see \cite{droniou2013gated} for exact definitions), the network is able to perform all computations directly in complex domain and neighboring weights in U now correspond to real and imaginary parts (see Fig. \ref{GAE_tied_fc_weights}). This way, the network is able to spare out $V$ of the real valued GAE (see also Figure \ref{model_architectures} c for a schematic depiction) which not only results in fewer parameters but also in a strong prior potentially speeding up convergence. While \cite{droniou2013gated} showed, how complex valued GAE can be used for reconstruction (see Eq. \ref{Droniou_reconstruction}), we were able to evidence that it is as capable for prediction:

\resizebox{.9\linewidth}{!}{
  \begin{minipage}{\linewidth}
\begin{align}
    X_t &= U'' R P (E_1 U'^T X_{t-1} \cdot E_3 W'^T\sigma(W P (E_1 U^T X_{t-1}\cdot E_2 U'^T X_t))) &\mathrm{reconstruction}\label{Droniou_reconstruction}\\
    X_{t+1} &= U'' R P (E_1 U'^T X_t \cdot E_3 W'^T\sigma(W P (E_1 U^T X_{t-1}\cdot E_2 U'^T X_t))) &\mathrm{prediction}
\end{align}
\end{minipage}
}

Since GAE scale quadratic in the number of image pixels, we also propose a convolutional form of the real and complex valued GAE which replaces the originally fully connected modules $U,V,W$ by convolutional units. The matrices $E_1,E_2,E_3,P,R$ for the complex valued GAE also stay the same but now are applied on the corresponding channels.

\ifx false
\newpage

\fi
\section{Experiments and Results}
\label{experimentsAndResults}
We performed several frame prediction experiments on different datasets in order to investigate the complex valued gated auto-encoder in the fully connected as well as in the convolutional setting. The network was trained to minimize the mean square error of the predicted frame with respect to the real follow-up frame.
\begin {table}[h]
\parbox{0.48\linewidth}{
\resizebox{0.75\linewidth}{!}{
\begin{minipage}{\linewidth}
\centering
\begin{tabular}{c c c}
\toprule
 & number of weights & Loss  \\
\midrule
real GAE, fc & 270400 & 0.63\\
complex GAE, fc & 155976 & 0.62\\
\bottomrule
\end{tabular}
\end{minipage}}
\caption{MSE on the Moving Noise}
\label{MN_mse}
}
\quad
\parbox{0.48\linewidth}{
\resizebox{0.6\linewidth}{!}{
  \begin{minipage}{\linewidth}
\centering
\begin{tabular}{c c c}
\toprule
 & number of weights & Loss  \\
\midrule
real GAE, fc & 1718400 & 3.3e-3 \\
complex GAE, fc & 903496 & 3.0e-3 \\
real GAE, conv & 5640 & 1.7e-4 \\
complex GAE, conv & 3241 & 2.1e-4 \\
real PGP, 2 layers, conv & 103240 & 3.8e-5 \\
complex PGP, 2 layers, conv & 52481 & 3.1e-5 \\
\bottomrule
\end{tabular}
\end{minipage}}
\caption{MSE on the Bouncing Ball}
\label{BB-shared-PGP}
}
\end{table}
\subsection{Moving Noise}
\begin{figure}[h]
\centering
\includegraphics[width=10cm]{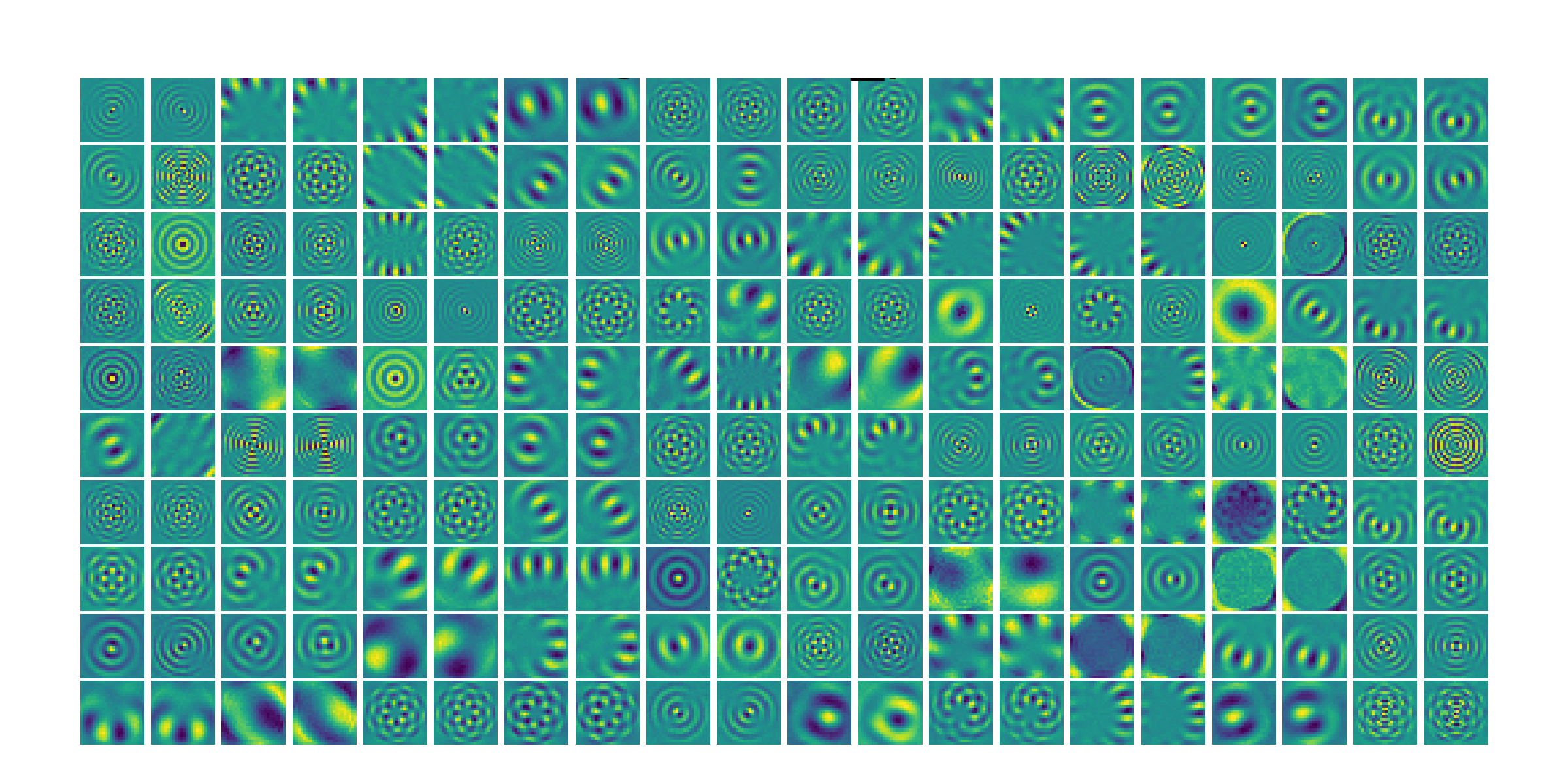}
\caption{weights of $U$ learned by the complex valued fully connected GAE. The pairing of neighboring real and imaginary parts is clearly visible.}
\label{GAE_tied_fc_weights}
\end{figure}

The moving noise dataset consists of image sequences containing Gaussian noise which is either uniformly moved in a random direction or uniformly rotated by a random angle. The resolution is 24 x 24 pixels. Table \ref{MN_mse} presents quantitative results and Figure \ref{GAE_tied_fc_weights} shows the weights in $U$ learned by the model after convergence. They visually look very similar to the weights obtained by \cite{droniou2013gated}, confirming that complex GAE can be used for prediction as well. Table \ref{MN_mse} shows that the complex valued GAE performs slightly better than the real valued GAE while incorporating significantly fewer weights.

\subsection{Bouncing Balls}
\begin{figure}[h]
\centering
\includegraphics[width=1\linewidth]{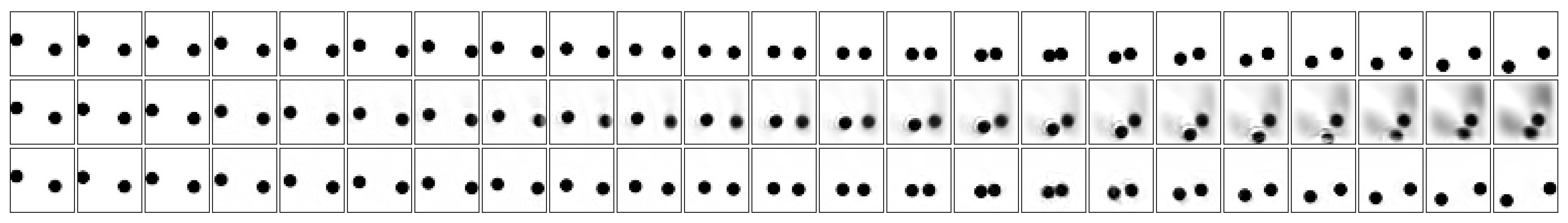}
\captionof{figure}{top row: ground truth, 2nd row: complex valued GAE, 3rd row: complex valued PGP. First 3 frames: seed frames, remaining 20 frames: predictions. In this example, the resolution is only 32 x 32 pixels and the training loss was averaged over 3 subsequent predicted frames to train the network on longer time horizons.}
\label{GAE_tied_conv_prediction}
\end{figure}

This dataset consists of 2 black balls that uniformly move in random directions of the 2D image plane. If a ball hits a wall or another ball, its movement gets reflected. The resolution is 64 x 64 pixels. Figure \ref{GAE_tied_conv_prediction} gives an idea about the qualitative results obtained using the convolutional complex valued GAE.

\begin{figure}[h]
\centering
\includegraphics[width=0.3\linewidth]{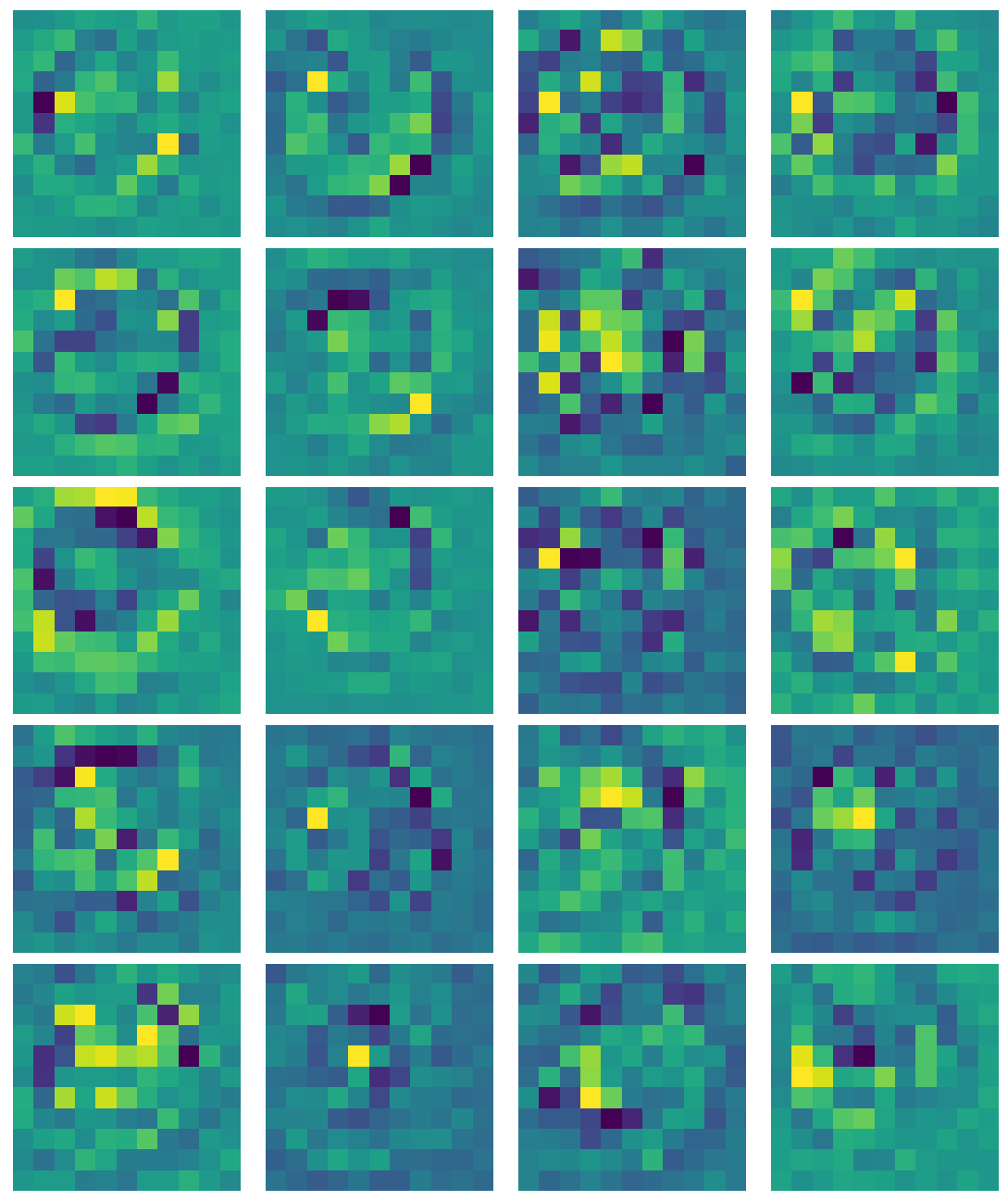}
\caption{weights of $U$ learned by the complex convolutional GAE}
\label{GAE_tied_conv_weights}
\end{figure}

Since both balls move in different directions, the space of possible transformations is huge. This makes the problem especially hard for a fully connected network. If we however use convolutional units whose kernels only cover a small part of the image, the transformations again correspond only to translations and can be more easily learned, while the number of parameters is drastically reduced. Our experiments (see Table \ref{BB-shared-PGP}) support this claim. Figure \ref{GAE_tied_conv_weights} clearly shows, that the convolutional model learns kernels that are able to shift a ball in different directions. Interactions between balls however cannot be modelled by this class of GAE as they violate the linearity assumption we made in the beginning:
\begin{equation}
    X_t^{\mathrm{ball}_1+\mathrm{ball}_2} = L(X_{t-1}^{\mathrm{ball}_1}+X_{t-1}^{\mathrm{ball}_2}) \neq LX_{t-1}^{\mathrm{ball}_1}+LX_{t-1}^{\mathrm{ball}_2} = X_t^{\mathrm{ball}_1}+X_t^{\mathrm{ball}_2}
\end{equation}
To deal with such cases, one has to also include higher order transformations - for example as shown by \citet{michalski2014modeling} with Predictive Gating Pyramids (PGP), which we refer to in the following as real valued PGP. A complex valued PGP can be obtained by replacing all real valued GAE inside the real valued PGP architecture by complex valued GAE. Experiments with real and complex valued PGP are also reported in Table \ref{BB-shared-PGP}. Qualitative results (see Figure \ref{GAE_tied_conv_prediction}) indeed show superior performance of the complex valued PGP over the complex valued GAE when interactions happen.

\section{Conclusion}
\label{conclusion}

In this work, we first showed, how the notoriously unstable deconvolution operation fits into the framework of gated auto-encoders. We then presented a way of extending complex valued GAE to perform predictions. Furthermore, we demonstrated that the complex convolutional form is more efficient than the complex fully connected form on the bouncing ball dataset. Our work puts some foundations on complex valued convolutional gated auto-encoders and closes the loop between real valued GAE \cite{memisevic2013learning} and complex valued GAE \cite{droniou2013gated} and PGP.

{\footnotesize \paragraph{Acknowledgment}
\label{acknowledgment}
This work was funded by grant BE 2556/16-1 (Research Unit FOR 2535 Anticipating Human Behavior) of the German Research Foundation (DFG).
}

\begin{footnotesize}
\bibliographystyle{unsrtnat}
\bibliography{references}

\end{footnotesize}

\end{document}